\documentclass[a4paper]{article}

\usepackage{INTERSPEECH2022}
\usepackage{multirow}
\usepackage{array}
\usepackage[table]{xcolor}

\newcolumntype{?}{!{\vrule width 1pt}}

\title{Investigation of Ensemble features of Self-Supervised Pretrained Models for Automatic Speech Recognition}
\name{A Arunkumar$^{*}$, Vrunda N Sukhadia$^{*}$, S. Umesh }
\address{Speech Lab, Dept. of Electrical Engineering, IIT Madras, Chennai, India}
\email{arunkumaras10@gmail.com, sukhadiavrunda@gmail.com, umeshs@ee.iitm.ac.in}

\begin{document}

\maketitle
\def\thefootnote{*}\footnotetext{Equal contribution}\def\thefootnote{\arabic{footnote}}
\begin{abstract}
Self-supervised learning (SSL) based models have been shown to generate powerful representations that can be used to improve the performance of downstream speech tasks. Several state-of-the-art SSL models are available, and each of these models optimizes a different loss which gives rise to the possibility of their features being complementary. This paper proposes using an ensemble of such SSL representations and models, which exploits the complementary nature of the features extracted by the various pretrained models. We hypothesize that this results in a richer feature representation and show results for the ASR downstream task. To this end, we use three SSL models that have shown excellent results on ASR tasks, namely HuBERT, Wav2vec2.0, and WaveLM. We explore the ensemble of models fine-tuned for the ASR task and the ensemble of features using the embeddings obtained from the pre-trained models for a downstream ASR task. We get improved performance over individual models and pre-trained features using Librispeech(100h) and WSJ dataset for the downstream tasks. \\
\end{abstract}


\section{Introduction}
Self-Supervised Learning (SSL) models \cite{oord2018representation, chung2019unsupervised, schneider2019wav2vec, baevski2019vq, liu2020mockingjay, baevski2020wav2vec, 9053569, 9478264, hsu2021hubert, chen2021wavlm} have been shown to provide significant improvement for various downstream tasks such as Automatic Speech Recognition(ASR), phoneme recognition, speaker verification, etc. Progress in ASR in many languages has been significantly affected due to the lack of good quality transcribed data. Self-supervised methods are highly desirable in such scenarios since they need a large amount of audio-only data, which is more readily available. 
Several pre-trained SSL models are now available in the public domain that can be used to build good downstream ASR models. These models vary in the loss function used during self-supervised learning and the type, nature, and amount of data used in self-supervision. We look at three models that have state-of-art ASR results for Libri subsets as well on SUPERB benchmarks \cite{baevski2020wav2vec,hsu2021hubert, chen2021wavlm,yang2021superb}, namely wav2vec2.0, HuBERT, and WavLM. We look at three models that have state-of-art ASR results for Libri subsets as well on SUPERB benchmarks \cite{baevski2020wav2vec,hsu2021hubert, chen2021wavlm,yang2021superb}, namely wav2vec2.0, HuBERT and WavLM. Each of these self-supervised training methods is motivated by a different objective:
\begin{itemize}
    \item Wave2vec2.0 masks latent representation and solves for a contrastive task with respect to quantised version of the latent representation
    \item HuBERT discovers acoustic-units using a clustering approach, which is used to label the input features. Masking is then applied on the input features, and the training is done to minimise the masked prediction loss using cluster labels as targets
    \item WavLM adds gated relative position bias to the transformer structure, and apart from masked prediction loss similar to HuBERT also applies denoising task during self-supervised learning. 
\end{itemize}
The last two models use non-contrastive criterion and therefore do not have to worry about large batch sizes which is important during training of wav2Vec2.0 models.  As different SSL models optimize different objective functions, each of them learns to extract different set of features. This opens the possibility of the extracted features to be complementary in nature. In this paper, an ensemble method which combines such complementary features from different pre-trained self-supervised models is investigated. Apart from the objective functions being different, each of these models have been trained on varying amounts of self-supervised data which are readily available in public domain. Similarly, many of these models are also available after fine-tuning for various tasks. 


The following state-of-the-art self-supervised models, which are readily available to download from HuggingFace \cite{wolf-etal-2020-transformers} are used in this paper:
\begin{itemize}
    \item "facebook/wav2vec2-base" trained on \textit{LibriSpeech} 960hrs data [95M parameters]
    \item "facebook/hubert-base-ls960" trained on \textit{LibriSpeech} 960hrs data [95M parameters]
    \item "microsoft/wavlm-base" trained on \textit{LibriSpeech} 960hrs data [94.70 parameters]
    \item "facebook/wav2vec2-large-lv60" trained on \textit{Libri-Light} 60k hours data [317.3M parameters]
    \item "facebook/hubert-xlarge-ll60k" trained on \textit{Libri-Light} 60k hours data [316.6M parameters]
    \item "microsoft/wavlm-large" trained on mix 94k hours data [60k hrs \textit{Libri-Light} + 10k hrs \textit{GigaSpeech} + 24k hrs \textit{VoxPopuli}] [316.6M parameters]
    \item Finetuned models "facebook/wav2vec2-large-960h-lv60-self" and "facebook/hubert-xlarge-ls960-ft" which are pre-trained on \textit{Libri-Light} 60k hours unlabeld data and finetuneed on \textit{LibriSpeech} 960 hours labeld data. 

\end{itemize}

\section{Fine Tuning for ASR task}
Pre-trained self-supervised models can be used in downstream ASR tasks in two ways:
\begin{itemize}
    \item We finetune the pre-trained self-supervised model using the supervised downstream dataset. This is done by adding a linear CTC layer \cite{graves2006connectionist} on top of the embeddings of the last layer output of the pre-traine model and optimising for the character output obtained from supervision. In this approach all the parameters of the pre-trained model are also updated (after freezing for a few iterations), and is usually expensive. For example, wav2vec2.0 fine-tuning on Libri splits take at least 50 V100 GPU hours \cite{baevski2020wav2vec}
    \item We freeze the pre-trained self-supervised model and use extracted features for downstream tasks. Since the SSL model parameters are not updated, a simple linear CTC layer on top during fine-tuning is not sufficient to get good results. In practice, couple of BiLSTM layers or transformer encoder layers are added on top of the pre-trained features followed by a CTC layers. During fine-tuning the BiLSTM/Transformer layers as well as the CTC layers are optmised for the supervised character output. Often a learnt linear combination of SSL layer outputs are used as features before feeding to the transformer encoder layers. 
\end{itemize}
This paper considers the strategy of freezing the pre-trained self-supervised model and using the models or extracted features for ASR tasks. Ensemble learning is a common approach in Deep Learning since it is predicated on the premise that combining the output of numerous models is more effective than using a single model and typically produces superior results. There are various approaches to combine features from multiple models including: 1) Summation of Features, 2) Weighted Average of Features, 3) Concatenation of Features, and 4) Soft mixing using Attention layer. In this paper, we  combine features from different models by concatenating extracted features and using the linear layer with CTC loss to learn the optimal combination of these concatenated features. However, since the SSL models are frozen, using only the CTC loss is not enough to guide best feature vector selection. Therefore, we also propose to use Transformer encoder \cite{vaswani2017attention} on top of the ensemble of features to compute a soft mixture of the features, this allows for even richer representation of features. A similar strategy is also adopted in S3PRL and SUPERB benchmarks. We first describe an approach to combine SSL models that have been specifically fine-tuned for ASR task. This is followed by a method to combine embeddings obtained from different pre-trained (but not fine-tuned) SLL models for a downstream ASR task.

\subsection{Ensemble Model}
\begin{figure}[h]
    \centering
    \includegraphics[width=\linewidth]{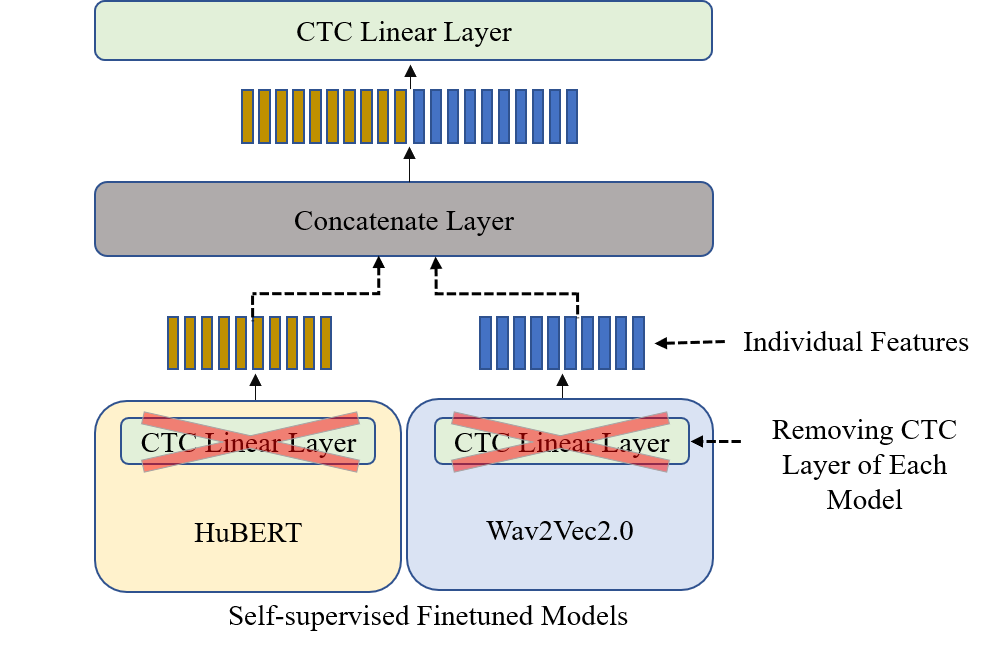}
    \caption{Proposed method for Ensemble model to combine HuBERT and wav2vec2.0 fine-tuned ASR models.}
    \label{fig:ensemble model}
\end{figure}

In this section, we describe our approach to combine SSL models that have been fine-tuned on ASR task. Specifically, we consider the wav2vec2.0 model fine-tuned on the Libri-960 hour supervised data, as well as the HuBERT model fine-tuned with the Libri-960 hour supervised data. These models have been fine-tuned with CTC Linear layer on top of the pre-trained transformer encoders with all the SSL model parameters being updated during fine-tuning (after being frozen for some initial iterations). In our proposed approach, we remove the final CTC layers from both the fine-tuned models and concatenate the two final layer embeddings. We now add a random CTC linear layer on top of these concatenated features and fine-tune on small amount of training data for a few epochs. Since these are already well fine-tuned models, the few epochs of fine-tuning help learn the concatenated features to character mapping. Please note that the fine-tuned SSL model parameters are \emph{not} updated, and only the CTC layer parameters on top of the concatenated features is learnt. This is shown in Figure~\ref{fig:ensemble model}. 

\begin{figure}[h]
    \centering
    \includegraphics[width=\linewidth]{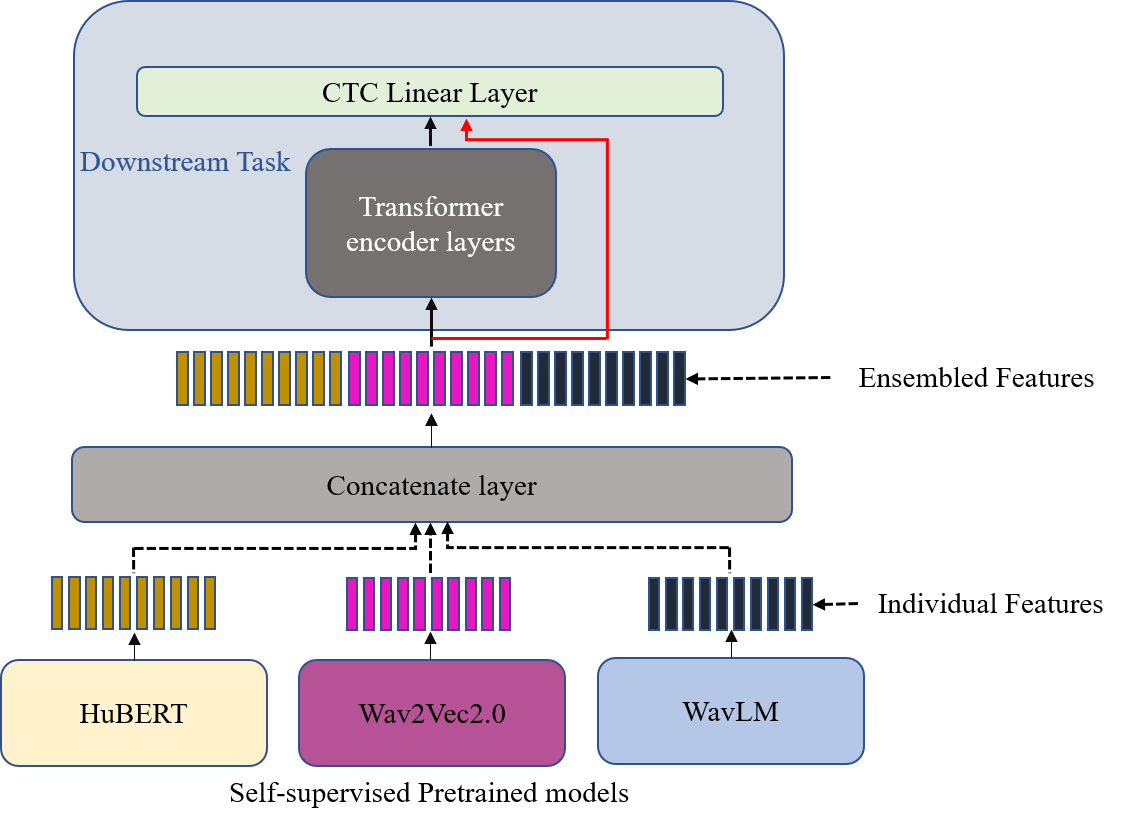}
    \caption{Proposed method to combine the embeddings from pre-trained models for downstream ASR task}
    \label{ensemble features}
\end{figure}

\subsection{Ensemble Features}
In this section, we describe how we combine the features or embeddings from the different pre-trained models for a downstream ASR task. Again, we freeze the pretrained the model parameters and do \emph{not} allow them to update. Therefore, a simple CTC layer on top of embeddings during fine-tuning will not give good performance. We need a few learnable layers on top to achieve good performance. This is similar to the approach taken by S3PRL and SUPERB benchmarks \cite{yang2021superb}, where a bi-directional LSTM on top of a pre-trained model is used for fine-tuning ASR downstream task. The authors proposed to freeze the pre-trained model and use weighted average of all the hidden layer features with trainable weights.  Instead of combining features from different layers of the same model, this work proposes to combine the last layer features of different pre-trained models and pass these features to transformer encoder layers.  We use transformer encoders instead of BiLSTM, since it offers faster processing with improved representation. The proposed ensemble is generated by combining multiple state-of-the-art models. All models in the ensemble are frozen, and feature representation collected from the final layer of each model is concatenated to produce an ensemble. This is shown in Figure~\ref{ensemble features}.


\section{Experiments}

\subsection{Ensemble Model}
For all the experiments presented in this section, large variants of wav2vec2 and HuBERT pre-trained with Libri-Light 60k hours and finetuned with LibriSpeech 960h are considered. These fine-tuned models are available for download in HuggingFace \cite{wolf-etal-2020-transformers}. All our experiments are done on the SpeechBrain toolkit \cite{ravanelli2021speechbrain}.  The CTC layer is removed from these models and the last layer features are extracted from these models. Note that these features are obtained from well fine-tuned ASR models, and therefore carry rich information about the final classification task of characters. While Wav2vec2.0 has 1024 dimension embeddings, HuBERT has 1280-dim embeddings. 
In our proposed approach, these features are concatenated before feeding it to a new randomized CTC layer.  The parameters of the linear classification layer are learnt using a few epochs on a small amount of training data. The linear layer helps learn appropriate combination of the concatenated features and a mapping of these features to characters.  To enable a fair comparison, the features from the individual models are obtained by removing the CTC layer of the fine-tuned model. These features are  fed to a randomised CTC layer whose parameters are also learnt by running a few epochs on a small amount of training data.

Table~\ref{ensemble model results} shows the result of combining the fine-tuned models using the approach of Figure~\ref{fig:ensemble model}. In this case, we have fine-tuned the randomised CTC layer using 10 hour of Libri data for about 8 epochs. The remaining model parameters are frozen. As seen from the table, using ensemble of models gives significant improvement over using individual models. Although both our models have been fine-tuned on 960-hour Librispeech, we also test them on WSJ data using the same approach. Once again we find improved performance using the proposed approach when compared to the individual models.
%

\begin{table}[h]
\centering
\footnotesize
\setlength\tabcolsep{3.5pt}
\begin{tabular}{c?c|c|c|c?c|c}
\hline
  \multirow{5}{*}{\textbf{Method}}  &
  \multicolumn{4}{c?}{\textbf{LibriSpeech -10h}} &
  \multicolumn{2}{c}{\textbf{WSJ -10h}} \\ \cline {2-7}
  &
  \multicolumn{2}{c|}{\textbf{clean}}  &
  \multicolumn{2}{c?}{\textbf{other}} &
  \multirow{3}{*}{\textbf{\begin{tabular}[c]{@{}c@{}}test\\ -Dev\\ 93\end{tabular}}} &
  \multirow{3}{*}{\textbf{\begin{tabular}[c]{@{}c@{}}test\\ -Eval\\ 93\end{tabular}}} \\
   \cline{2-5}
  &
  \multirow{2}{*}{\textbf{dev}}&
  \multirow{2}{*}{\textbf{test}}&
  \multirow{2}{*}{\textbf{dev}}&
  \multirow{2}{*}{\textbf{test}}&
  &
  \\
  &
  &
  &
  &
  &
  \\
  \hline\hline
\begin{tabular}[c]{@{}c@{}}Wav2Vec\\ 2.0\end{tabular} &
  \multicolumn{1}{c|}{4.13} &
  \multicolumn{1}{c|}{4.31} &
  \multicolumn{1}{c|}{6.91} &
  7.21 &
  \multicolumn{1}{c|}{37.21} &
  35.36 \\ \hline
HuBERT &
  \multicolumn{1}{c|}{4.35} &
  \multicolumn{1}{c|}{4.03} &
  \multicolumn{1}{c|}{5.79} &
  6.30 &
  \multicolumn{1}{c|}{7.32} &
   6.54\\ \hline
\begin{tabular}[c]{@{}c@{}}Wav2Vec\\ 2.0 \\ + HuBERT\end{tabular} &
  \multicolumn{1}{c|}{3.15} &
  \multicolumn{1}{c|}{3.05} &
  \multicolumn{1}{c|}{4.76} &
  5.16 &
  \multicolumn{1}{c|}{6.79} &
  6.13 \\ \hline
\end{tabular}
\caption{Evaluating Results Of Individual SSL Methods And Ensemble Model For In-domain (\textit{Libri-Light} 10h) And Out-domain (10h Subset Of \textit{WSJ}) Downstream Data. Only CTC Layer Is Added On Top Of Features. No External Language Model is Used.}
\label{ensemble model results}
\end{table}

\subsection{Ensemble Features}
In this section, we present results of using ensemble of features from pre-trained models to improve the performance of the downstream ASR task.
Both the base and large variants of pre-trained models are used for the ensemble feature experiments. In this section apart from Wav2vec2.0 and HuBERT, we also consider the pre-trained WavLM model since this is also readily available in HuggingFace \cite{wolf-etal-2020-transformers}. All these are experiments were implemented in speechbrain toolkit \cite{ravanelli2021speechbrain}.  The last layer features are extracted from pre-trained model and then passed to the CTC layer or transformer network. Note that all the prertrained models we are using have been trained on Librispeech data in a self-supervised manner. We now conduct experiments in two scenarios: (i) where the fine-tuning is done on matched data, namely Libri subsets, and (ii) where the fine-tuning is done a different domain data, namely WSJ. Since these embeddings are obtained from pre-trained models that have not been tuned for ASR task, and since we do not allow the pre-trained model parameters to be updated, a simple CTC linear on top does not give good performance. We need additional learnable layers to get good performance, and in our case we use transformer encoder layers.
\subsubsection{Finetuning with Same Domain Data}
In this section, the pre-trained models are fine-tuned with data from same domain, namely, Libri-100 hour data.

\begin{table}[h]
\setlength\tabcolsep{2pt}
\footnotesize
\begin{tabular}{c?cccccccc}
\hline\hline
  \multirow{3}{*}{\textbf{\begin{tabular}[c]{@{}c@{}}Feature\\ Extraction\\ Model\end{tabular}}} &
  \multicolumn{4}{c?}{\textbf{CTC Layer only}} &
  \multicolumn{4}{c}{\textbf{2 encoder layers + CTC}} \\ \cline{2-9}
 &
  \multicolumn{2}{c|}{\textbf{Clean}} &
  \multicolumn{2}{c?}{\textbf{Other}} &
  \multicolumn{2}{c|}{\textbf{Clean}} &
  \multicolumn{2}{c}{\textbf{Other}} \\ \cline{2-9} 
 &
  \multicolumn{1}{c|}{\textbf{dev}} &
  \multicolumn{1}{c|}{\textbf{test}} &
  \multicolumn{1}{c|}{\textbf{dev}} &
  \multicolumn{1}{c?}{\textbf{test}} &
  \multicolumn{1}{c|}{\textbf{dev}} &
  \multicolumn{1}{c|}{\textbf{test}} &
  \multicolumn{1}{c|}{\textbf{dev}} &
  test \\ \hline \hline
\multicolumn{9}{c}{Base models} \\ \hline
\multicolumn{1}{c?}{\begin{tabular}[c]{@{}c@{}}WavLM\\ {}\end{tabular}} &
  \multicolumn{1}{c|}{59.54} &
  \multicolumn{1}{c|}{60.19} &
  \multicolumn{1}{c|}{67.73} &
  \multicolumn{1}{c?}{67.85} &
  \multicolumn{1}{c|}{9.65} &
  \multicolumn{1}{c|}{10.34} &
  \multicolumn{1}{c|}{20.87} &
  20.88 \\ \hline
\multicolumn{1}{c?}{\begin{tabular}[c]{@{}c@{}}HuBERT\\ {}\end{tabular}} &
  \multicolumn{1}{c|}{67.96} &
  \multicolumn{1}{c|}{68.43} &
  \multicolumn{1}{c|}{76.33} &
  \multicolumn{1}{c?}{76.2} &
  \multicolumn{1}{c|}{12.41} &
  \multicolumn{1}{c|}{13.37} &
  \multicolumn{1}{c|}{24.8} &
  25.16 \\ \hline
  \multicolumn{1}{c?}{\begin{tabular}[c]{@{}c@{}}Wav2Vec2.0\\ {}\end{tabular}} &
  \multicolumn{1}{c|}{98.84} &
  \multicolumn{1}{c|}{98.54} &
  \multicolumn{1}{c|}{99.16} &
  \multicolumn{1}{c?}{98.97} &
  \multicolumn{1}{c|}{28.56} &
  \multicolumn{1}{c|}{29.68} &
  \multicolumn{1}{c|}{44.99} &
  46.05 \\ \hline
\multicolumn{1}{c?}{\begin{tabular}[c]{@{}c@{}}WavLM\\ +HuBERT\end{tabular}} &
  \multicolumn{1}{c|}{51.02} &
  \multicolumn{1}{c|}{51.57} &
  \multicolumn{1}{c|}{60.43} &
  \multicolumn{1}{c?}{60.59} &
  \multicolumn{1}{c|}{8.72} &
  \multicolumn{1}{c|}{9.25} &
  \multicolumn{1}{c|}{19.63} &
  19.60 \\ \hline
\multicolumn{1}{c?}{\begin{tabular}[c]{@{}c@{}}WavLM\\ +Wav2Vec2.0\end{tabular}} &
  \multicolumn{1}{c|}{60.08} &
  \multicolumn{1}{c|}{60.64} &
  \multicolumn{1}{c|}{68.43} &
  \multicolumn{1}{c?}{68.57} &
  \multicolumn{1}{c|}{11.58} &
  \multicolumn{1}{c|}{12.59} &
  \multicolumn{1}{c|}{23.38} &
  23.60 \\ \hline
\multicolumn{1}{c?}{\begin{tabular}[c]{@{}c@{}}Wav2Vec2.0\\ +HuBERT\end{tabular}} &
  \multicolumn{1}{c|}{68.59} &
  \multicolumn{1}{c|}{68.49} &
  \multicolumn{1}{c|}{76.95} &
  \multicolumn{1}{c?}{76.19} &
  \multicolumn{1}{c|}{13.77} &
  \multicolumn{1}{c|}{14.34} &
  \multicolumn{1}{c|}{26.19} &
  27.0 \\ \hline
\multicolumn{1}{c?}{\begin{tabular}[c]{@{}c@{}}Wav2Vec2.0\\ +HuBERT\\ +WavLM\end{tabular}} &
  \multicolumn{1}{c|}{52.21} &
  \multicolumn{1}{c|}{52.87} &
  \multicolumn{1}{c|}{61.53} &
  \multicolumn{1}{c?}{61.73} &
  \multicolumn{1}{c|}{9.03} &
  \multicolumn{1}{c|}{9.58} &
  \multicolumn{1}{c|}{19.96} &
  20.16 \\ \hline
\multicolumn{9}{c}{Large models} \\ \hline
\multicolumn{1}{c?}{\begin{tabular}[c]{@{}c@{}}WavLM\\ {}\end{tabular}} &
  \multicolumn{1}{c|}{47.02} &
  \multicolumn{1}{c|}{47.66} &
  \multicolumn{1}{c|}{54.04} &
  \multicolumn{1}{c?}{53.78} &
  \multicolumn{1}{c|}{5.79} &
  \multicolumn{1}{c|}{5.97} &
  \multicolumn{1}{c|}{11.09} &
   11.01\\ \hline
\multicolumn{1}{c?}{\begin{tabular}[c]{@{}c@{}}HuBERT\\ {}\end{tabular}} &
  \multicolumn{1}{c|}{58.98} &
  \multicolumn{1}{c|}{58.44} &
  \multicolumn{1}{c|}{62.64} &
  \multicolumn{1}{c?}{62.19} &
  \multicolumn{1}{c|}{9.94} &
  \multicolumn{1}{c|}{10.36} &
  \multicolumn{1}{c|}{13.54} &
   13.93\\ \hline
\multicolumn{1}{c?}{\begin{tabular}[c]{@{}c@{}}Wav2Vec2.0\\ {}\end{tabular}} &
  \multicolumn{1}{c|}{99.99} &
  \multicolumn{1}{c|}{99.99} &
  \multicolumn{1}{c|}{99.97} &
  \multicolumn{1}{c?}{99.98} &
  \multicolumn{1}{c|}{98.59} &
  \multicolumn{1}{c|}{98.57} &
  \multicolumn{1}{c|}{98.67} &
   98.64\\ \hline
\multicolumn{1}{c?}{\begin{tabular}[c]{@{}c@{}}WavLM\\ +HuBERT\end{tabular}} &
  \multicolumn{1}{c|}{41.17} &
  \multicolumn{1}{c|}{41.61} &
  \multicolumn{1}{c|}{46.38} &
  \multicolumn{1}{c?}{46.06} &
  \multicolumn{1}{c|}{5.60} &
  \multicolumn{1}{c|}{5.49} &
  \multicolumn{1}{c|}{9.76} &
   9.53\\ \hline
\multicolumn{1}{c?}{\begin{tabular}[c]{@{}c@{}}WavLM\\ +Wav2Vec2.0\end{tabular}} &
  \multicolumn{1}{c|}{47.44} &
  \multicolumn{1}{c|}{48.01} &
  \multicolumn{1}{c|}{53.94} &
  \multicolumn{1}{c?}{53.89} &
  \multicolumn{1}{c|}{6.64} &
  \multicolumn{1}{c|}{6.99} &
  \multicolumn{1}{c|}{12.22} &
   12.21\\ \hline
\multicolumn{1}{c?}{\begin{tabular}[c]{@{}c@{}}Wav2Vec2.0\\ +HuBERT\end{tabular}} &
  \multicolumn{1}{c|}{56.47} &
  \multicolumn{1}{c|}{55.84} &
  \multicolumn{1}{c|}{66.77} &
  \multicolumn{1}{c?}{60.69} &
  \multicolumn{1}{c|}{9.64} &
  \multicolumn{1}{c|}{9.90} &
  \multicolumn{1}{c|}{12.96} &
   13.35\\ \hline
\multicolumn{1}{c?}{\begin{tabular}[c]{@{}c@{}}Wav2Vec2.0\\ +HuBERT\\ +WavLM\end{tabular}} &
  \multicolumn{1}{c|}{40.00} &
  \multicolumn{1}{c|}{40.24} &
  \multicolumn{1}{c|}{46.21} &
  \multicolumn{1}{c?}{46.37} &
  \multicolumn{1}{c|}{} &
  \multicolumn{1}{c|}{} &
  \multicolumn{1}{c|}{} &
   \\ \hline
\end{tabular}
\caption{}
\label{Librispeech results}
\end{table}

A CTC only model or a downstream transformer encoder model with CTC is finetuned with the individual features and the ensemble features extracted from LibriSpeech 100h data. In the first set of experiments, only a CTC layer is finetuned. In the second set of experiments, two layer transformer encoder with a CTC layer on top is finetuned. The results are shown in Table \ref{Librispeech results}.  Since, the features are extracted from a pre-trained model and not from a finetuned model, CTC only finetuning is not good enough. It can be seen that the ensemble features model improves over the individual features model. Even though the performance is poor in case of CTC only finetuning, the ensemble features is still relatively better than the individual features. The more interesting results are with the use of transformer encoder layers trained on top of these features. From Table~\ref{Librispeech results}, the following observations can be made:
\begin{itemize}
    \item SSL Models trained on larger amounts of data consistently gave better perform on the same downstream task as expected.
    \item WavLM consistently performs better than HuBERT which is better than Wav2Vec2.0 in all cases for this ASR task.
    \item Wav2vec2.0 features are significantly worse, and therefore hurt performance when combined with other features. This is because as observed in previous works, the final layer of WavLM and HuBERT capture significant layer for ASR task, while it is the middle layers of Wav2Vec2.0 that are more appropriate for ASR task.
    \item Combining WavLM and HuBERT features gives the best performance for this task, and significantly better than the individual models. 
    \item Compare to the best individual model performance, the WavLM+HuBERT features give a relative 10\% improvement.
\end{itemize}
\subsubsection{Finetuning with Different Domain Data}
In this section, we fine-tune with WSJ data for the downstream ASR task. While WSJ data is mostly from business domain with modern English, Librispeech is mostly audiobooks of old English material from project Gutenberg. Therefore, there is a mismatch in domain. Since the domain of labeled data used for finetuning is different from the data used for pre-training, we need a more complex model to learn better. Therefore, while for the previous section we can got good performance for 2-layer encoder, for the mismatched WSJ task we have used 8 encoder layers followed by CTC to get good performance. In the first set of experiments, only a CTC layer is applied on top and finetuned. As expected these perform far worse in this mismatched case. In the next set of experiments, eight layer transformer encoder is finetuned. From Table~\ref{WSJ results} we make the following observations:
\begin{itemize}
    \item For fine-tuning models from a different domain too, the WavLM model performs better than HuBERT which is better than Wav2Vec2.0
    \item In this case also, a combination of WavLM+HuBERT gives the best performance.
    \item The relative improvement using ensemble features is more significant when using large pretrained models when compared to the base models.
    \item Even for the WSJ task, the ensemble feature provide about 10\% relative improvement over the best performing individual features.
\end{itemize}

\begin{table}[h]
\footnotesize
\setlength\tabcolsep{2.5pt}
\begin{tabular}{c?c|c?c|c}
\hline \hline
  \multirow{3}{*}{\textbf{\begin{tabular}[c]{@{}c@{}}Feature\\ Extraction\\ Model\end{tabular}}} &
  \multicolumn{2}{c?}{\textbf{CTC Layer only}} &
  \multicolumn{2}{c}{\textbf{8 encoder layers + CTC}} \\ \cline{2-5}
\multicolumn{1}{c?}{} &
  \multirow{2}{*}{\textbf{Test-Dev9}3} & 
  \multirow{2}{*}{\textbf{Test-Eval93}} &
  \multirow{2}{*}{\textbf{Test-Dev93}} &
  \multirow{2}{*}{\textbf{Test-Eval93}} \\ 
\multicolumn{1}{c?}{} &
\multicolumn{1}{c|}{} & 
\multicolumn{1}{c?}{} &
\multicolumn{1}{c|}{} &
\multicolumn{1}{c}{} \\\hline\hline
\multicolumn{5}{c}{Base models} \\ \hline
\multicolumn{1}{c?}{\begin{tabular}[c]{@{}c@{}}WavLM\\ {}\end{tabular}} &
  \multicolumn{1}{c|}{66.81} &
  \multicolumn{1}{c?}{65.28} &
  \multicolumn{1}{c|}{15.77} &
  14.93 \\ \hline
\multicolumn{1}{c?}{\begin{tabular}[c]{@{}c@{}}HuBERT\\ {}\end{tabular}} &
  \multicolumn{1}{c|}{74.97} &
  \multicolumn{1}{c?}{75.01} &
  \multicolumn{1}{c|}{22.54} &
  20.69 \\ \hline
\multicolumn{1}{c?}{\begin{tabular}[c]{@{}c@{}}Wav2Vec2.0\\ {}\end{tabular}} &
  \multicolumn{1}{c|}{} &
  \multicolumn{1}{c?}{} &
  \multicolumn{1}{c|}{35.12} &
  31.84 \\ \hline
\multicolumn{1}{c?}{\begin{tabular}[c]{@{}c@{}}WavLM\\ +HuBERT\end{tabular}} &
  \multicolumn{1}{c|}{59.46} &
  \multicolumn{1}{c?}{58.05} &
  \multicolumn{1}{c|}{16.04} &
  14.93 \\ \hline
\multicolumn{1}{c?}{\begin{tabular}[c]{@{}c@{}}WavLM\\ +Wav2Vec2.0\end{tabular}} &
  \multicolumn{1}{c|}{68.05} &
  \multicolumn{1}{c?}{67.37} &
  \multicolumn{1}{c|}{16.24} &
  15.37 \\ \hline
\multicolumn{1}{c?}{\begin{tabular}[c]{@{}c@{}}Wav2Vec2.0\\ +HuBERT\end{tabular}} &
  \multicolumn{1}{c|}{74.91} &
  \multicolumn{1}{c?}{74.87} &
  \multicolumn{1}{c|}{17.94} &
  17.66 \\ \hline
\multicolumn{1}{c?}{\begin{tabular}[c]{@{}c@{}}Wav2Vec2.0\\ +HuBERT\\ +WavLM\end{tabular}} &
  \multicolumn{1}{c|}{60.14} &
  \multicolumn{1}{c?}{59.79} &
  \multicolumn{1}{c|}{16.4} &
  15.43 \\ \hline
\multicolumn{5}{c}{Large models} \\ \hline
\multicolumn{1}{c?}{\begin{tabular}[c]{@{}c@{}}WavLM\\ {} \end{tabular}} &
  \multicolumn{1}{c|}{55.83} &
  \multicolumn{1}{c?}{55.08} &
  \multicolumn{1}{c|}{11.09} &
  10.37 \\ \hline
\multicolumn{1}{c?}{\begin{tabular}[c]{@{}c@{}}HuBERT\\ {} \end{tabular}} &
  \multicolumn{1}{c|}{66.97} &
  \multicolumn{1}{c?}{66.47} &
  \multicolumn{1}{c|}{13.22} &
  12.73 \\ \hline
\multicolumn{1}{c?}{\begin{tabular}[c]{@{}c@{}}Wav2Vec2.0\\ {} \end{tabular}} &
  \multicolumn{1}{c|}{98.01} &
  \multicolumn{1}{c?}{98.08} &
  \multicolumn{1}{c|}{} &
  \multicolumn{1}{l}{} \\ \hline
\multicolumn{1}{c?}{\begin{tabular}[c]{@{}c@{}}WavLM\\ +HuBERT\end{tabular}} &
  \multicolumn{1}{c|}{49.81} &
  \multicolumn{1}{c?}{49.30} &
  \multicolumn{1}{c|}{9.57} &
  8.86 \\ \hline
\multicolumn{1}{c?}{\begin{tabular}[c]{@{}c@{}}WavLM\\ +Wav2Vec2.0\end{tabular}} &
  \multicolumn{1}{c|}{56.80} &
  \multicolumn{1}{c?}{56.13} &
  \multicolumn{1}{c|}{10.32} &
  9.59 \\ \hline
\multicolumn{1}{c?}{\begin{tabular}[c]{@{}c@{}}Wav2Vec2.0\\ +HuBERT\end{tabular}} &
  \multicolumn{1}{c|}{66.68} &
  \multicolumn{1}{c?}{65.43} &
  \multicolumn{1}{c|}{18.50} &
  17.75 \\ \hline
\multicolumn{1}{c?}{\begin{tabular}[c]{@{}c@{}}Wav2Vec2.0\\ +HuBERT\\ +WavLM\end{tabular}} &
  \multicolumn{1}{c|}{48.79} &
  \multicolumn{1}{c?}{47.62} &
  \multicolumn{1}{c|}{15.09} &
  13.89 \\ \hline
\end{tabular}
\caption{Evaluating Results Of Individual Ssl Methods And Ensemble Model For In-domain (\textit{Libri-Light} 10h) And Out-domain (10h Subset Of WSJ) Downstream Data. Only Ctc Layer Is Added On Top Of Features. No External Language Model Is Used.}
\label{WSJ results}
\end{table}

\section{Conclusion}
In this paper, we have explored the use of ensemble of models and embedding from pre-trained model to improve the performance over individual SSL methods. In all cases, we have used publicly available models. The motivation for the use of ensemble is that different SSL methods employ different objective functions such masked prediction loss or contrastive loss. Therefore, they may capture complimentary information. On the downstream ASR, the use of our proposed approaches provide a relative improvement of 10\% over the best individual models for both Libri-100 as well as WSJ task. 

\section{Acknowledgement}
We would like to thank Metilda for technical discussion and all her help in preparing this paper.

\bibliographystyle{IEEEtran}
\bibliography{main}

\begin{thebibliography}{10}
\providecommand{\url}[1]{#1}
\csname url@samestyle\endcsname
\providecommand{\newblock}{\relax}
\providecommand{\bibinfo}[2]{#2}
\providecommand{\BIBentrySTDinterwordspacing}{\spaceskip=0pt\relax}
\providecommand{\BIBentryALTinterwordstretchfactor}{4}
\providecommand{\BIBentryALTinterwordspacing}{\spaceskip=\fontdimen2\font plus
\BIBentryALTinterwordstretchfactor\fontdimen3\font minus
  \fontdimen4\font\relax}
\providecommand{\BIBforeignlanguage}[2]{{%
\expandafter\ifx\csname l@#1\endcsname\relax
\typeout{** WARNING: IEEEtran.bst: No hyphenation pattern has been}%
\typeout{** loaded for the language `#1'. Using the pattern for}%
\typeout{** the default language instead.}%
\else
\language=\csname l@#1\endcsname
\fi
#2}}
\providecommand{\BIBdecl}{\relax}
\BIBdecl

\bibitem{oord2018representation}
A.~v.~d. Oord, Y.~Li, and O.~Vinyals, ``Representation learning with
  contrastive predictive coding,'' \emph{arXiv preprint arXiv:1807.03748},
  2018.

\bibitem{chung2019unsupervised}
Y.-A. Chung, W.-N. Hsu, H.~Tang, and J.~Glass, ``An unsupervised autoregressive
  model for speech representation learning,'' \emph{arXiv preprint
  arXiv:1904.03240}, 2019.

\bibitem{schneider2019wav2vec}
S.~Schneider, A.~Baevski, R.~Collobert, and M.~Auli, ``wav2vec: Unsupervised
  pre-training for speech recognition,'' \emph{arXiv preprint
  arXiv:1904.05862}, 2019.

\bibitem{baevski2019vq}
A.~Baevski, S.~Schneider, and M.~Auli, ``vq-wav2vec: Self-supervised learning
  of discrete speech representations,'' \emph{arXiv preprint arXiv:1910.05453},
  2019.

\bibitem{liu2020mockingjay}
A.~T. Liu, S.-w. Yang, P.-H. Chi, P.-c. Hsu, and H.-y. Lee, ``Mockingjay:
  Unsupervised speech representation learning with deep bidirectional
  transformer encoders,'' in \emph{ICASSP 2020-2020 IEEE International
  Conference on Acoustics, Speech and Signal Processing (ICASSP)}.\hskip 1em
  plus 0.5em minus 0.4em\relax IEEE, 2020, pp. 6419--6423.

\bibitem{baevski2020wav2vec}
A.~Baevski, Y.~Zhou, A.~Mohamed, and M.~Auli, ``wav2vec 2.0: A framework for
  self-supervised learning of speech representations,'' \emph{Advances in
  Neural Information Processing Systems}, vol.~33, pp. 12\,449--12\,460, 2020.

\bibitem{9053569}
M.~Ravanelli, J.~Zhong, S.~Pascual, P.~Swietojanski, J.~Monteiro, J.~Trmal, and
  Y.~Bengio, ``Multi-task self-supervised learning for robust speech
  recognition,'' in \emph{ICASSP 2020 - 2020 IEEE International Conference on
  Acoustics, Speech and Signal Processing (ICASSP)}, 2020, pp. 6989--6993.

\bibitem{9478264}
A.~T. Liu, S.-W. Li, and H.-y. Lee, ``Tera: Self-supervised learning of
  transformer encoder representation for speech,'' \emph{IEEE/ACM Transactions
  on Audio, Speech, and Language Processing}, vol.~29, pp. 2351--2366, 2021.

\bibitem{hsu2021hubert}
W.-N. Hsu, B.~Bolte, Y.-H.~H. Tsai, K.~Lakhotia, R.~Salakhutdinov, and
  A.~Mohamed, ``Hubert: Self-supervised speech representation learning by
  masked prediction of hidden units,'' \emph{IEEE/ACM Transactions on Audio,
  Speech, and Language Processing}, vol.~29, pp. 3451--3460, 2021.

\bibitem{chen2021wavlm}
S.~Chen, C.~Wang, Z.~Chen, Y.~Wu, S.~Liu, Z.~Chen, J.~Li, N.~Kanda,
  T.~Yoshioka, X.~Xiao \emph{et~al.}, ``Wavlm: Large-scale self-supervised
  pre-training for full stack speech processing,'' \emph{arXiv preprint
  arXiv:2110.13900}, 2021.

\bibitem{yang2021superb}
S.-w. Yang, P.-H. Chi, Y.-S. Chuang, C.-I.~J. Lai, K.~Lakhotia, Y.~Y. Lin,
  A.~T. Liu, J.~Shi, X.~Chang, G.-T. Lin \emph{et~al.}, ``Superb: Speech
  processing universal performance benchmark,'' \emph{arXiv preprint
  arXiv:2105.01051}, 2021.

\bibitem{wolf-etal-2020-transformers}
\BIBentryALTinterwordspacing
T.~Wolf, L.~Debut, V.~Sanh, J.~Chaumond, C.~Delangue, A.~Moi, P.~Cistac,
  T.~Rault, R.~Louf, M.~Funtowicz, J.~Davison, S.~Shleifer, P.~von Platen,
  C.~Ma, Y.~Jernite, J.~Plu, C.~Xu, T.~Le~Scao, S.~Gugger, M.~Drame, Q.~Lhoest,
  and A.~Rush, ``Transformers: State-of-the-art natural language processing,''
  in \emph{Proceedings of the 2020 Conference on Empirical Methods in Natural
  Language Processing: System Demonstrations}.\hskip 1em plus 0.5em minus
  0.4em\relax Online: Association for Computational Linguistics, Oct. 2020, pp.
  38--45. [Online]. Available:
  \url{https://aclanthology.org/2020.emnlp-demos.6}
\BIBentrySTDinterwordspacing

\bibitem{graves2006connectionist}
A.~Graves, S.~Fern{\'a}ndez, F.~Gomez, and J.~Schmidhuber, ``Connectionist
  temporal classification: labelling unsegmented sequence data with recurrent
  neural networks,'' in \emph{Proceedings of the 23rd international conference
  on Machine learning}, 2006, pp. 369--376.

\bibitem{vaswani2017attention}
A.~Vaswani, N.~Shazeer, N.~Parmar, J.~Uszkoreit, L.~Jones, A.~N. Gomez,
  {\L}.~Kaiser, and I.~Polosukhin, ``Attention is all you need,''
  \emph{Advances in neural information processing systems}, vol.~30, 2017.

\bibitem{ravanelli2021speechbrain}
M.~Ravanelli, T.~Parcollet, P.~Plantinga, A.~Rouhe, S.~Cornell, L.~Lugosch,
  C.~Subakan, N.~Dawalatabad, A.~Heba, J.~Zhong \emph{et~al.}, ``Speechbrain: A
  general-purpose speech toolkit,'' \emph{arXiv preprint arXiv:2106.04624},
  2021.

\end{thebibliography}

\end{document}